\documentclass{article}

\usepackage[preprint,nonatbib]{neurips_2023}

\usepackage[utf8]{inputenc} 
\usepackage[T1]{fontenc}    
\usepackage{hyperref}       
\usepackage{url}            
\usepackage{booktabs}       
\usepackage{amsfonts}       
\usepackage{nicefrac}       
\usepackage{microtype}      
\usepackage{xcolor}         

\usepackage{graphicx}
\usepackage{amsmath}
\usepackage{subfig} 
\usepackage{pythonhighlight}
\usepackage{biblatex}
\usepackage{booktabs}
\usepackage{bm}
\usepackage{enumerate}
\title{Optimal and Efficient Binary Questioning for Human-in-the-Loop Annotation}

\author{%
  Franco Marchesoni-Acland\\
  SLB, Centre Borelli\\
  \texttt{marchesoniacland@gmail.com} \\
  \And
  Jean-Michel Morel \\
  Centre Borelli \\
  Université Paris-Saclay\\
  \AND
  Josselin Kherroubi\\
  SLB\\
  \texttt{JKherroubi@slb.com}
  \And
  Gabriele Facciolo\\
  Centre Borelli\\
  Université Paris-Saclay\\
}

\begin{document}

\maketitle

\begin{abstract}
Even though data annotation is extremely important for interpretability, research and development of artificial intelligence solutions, most research efforts such as active learning or few-shot learning focus on the sample efficiency problem.
This paper studies the neglected complementary problem of getting annotated data given a predictor. For the simple binary classification setting, we present the spectrum ranging from optimal general solutions to practical efficient methods. The problem is framed as the full annotation of a binary classification dataset with the minimal number of yes/no questions when a predictor is available. For the case of general binary questions the solution is found in coding theory, where the optimal questioning strategy is given by the Huffman encoding of the possible labelings. However, this approach is computationally intractable even for small dataset sizes.
We propose an alternative practical solution based on several heuristics and lookahead minimization of proxy cost functions.
The proposed solution is analysed, compared with optimal solutions and evaluated on several synthetic and real-world datasets. On these datasets, the method allows a significant improvement ($23-86\%$) in annotation efficiency.
\end{abstract}



\section{Introduction}

Deep supervised learning has been key to advancing artificial intelligence (AI) and solving practical problems. Even now, with the advent of more powerful self-supervised methods, human feedback is essential to ensure proper deployment and correct downstream performance \cite{bai2022training-rlhf, oquab2023dinov2}. Annotated datasets help researchers evaluate performances on tasks that can be as subjective as image enhancement \cite{fivek-mit5k} and allow engineers to fine-tune self-supervised predictors on downstream tasks. Although rarely perfect, datasets are fundamental for AI research and development, yet the quick annotation of datasets is still an open problem.

There is a strong relation between data and (trained) predictors. A large portion of the last decade of research focused on supervised learning, i.e. on obtaining better predictors given some annotated data. More recently the research evolved and refocused into sampling efficiency considerations, as seen in active learning (AL), few-shot learning (FSL) or semi-supervised learning. Alternatively, data availability was considered by the self-supervised learning or weakly-supervised learning fields. Virtually all these fields study the path from data to predictors. In contrast, there have been very few works (e.g. \cite{NEURIPS2020_05f971b5-onebit}) studying the path from predictors to data. This opposite direction lies in the core of quick or intelligent annotation (IA) solutions, i.e. methods that allow a human to quickly annotate a dataset. Most of these methods have a predictor of some sort or could benefit from one.

From a higher level, human-in-the-loop learning (HILL) comprises methods that involve the interaction between a human and a machine with a teaching/learning aim. The most basic way for a human to teach a machine a task is not HILL but is still widely used: the human simply annotates a dataset and a predictor is trained on it. More sophisticated methods involve annotating the dataset by leveraging a fixed generalistic predictor, as it is done in interactive image segmentation \cite{kirillov2023segment-sam, liu2023simpleclick}. In the case where the stopping criteria for annotation is a performance threshold the HILL solution of active learning allows reducing the annotation effort by stopping the annotation earlier. What has been largely missing from these solutions is using the predictor to reduce the annotation cost. Leveraging a predictor has been shown to increase annotation efficacy in practice \cite{NEURIPS2020_05f971b5-onebit} and in theory \cite{kulkarni1993active-jm1, yang2010bayesian-jm2}.

The importance of data annotation follows from the importance of annotated data.
Even though there are some problems that are truly self-supervised in the sense of not requiring any human supervision (e.g. generative text models \cite{yang2023harnessing-gtmreview}), many of those solutions require some supervision for finetuning \cite{oquab2023dinov2, liu2023simpleclick} or alignment \cite{bai2022training-rlhf}. Of course, annotated data is not only used for development but also for testing and comparing predictors. 

The present work analyses the quick annotation problem for the binary classification setting. The quick annotation problem involves finding the method that annotates more samples with the least number of yes/no answers. The direction goes from predictor to data, opposite to the usual data-to-predictor direction. Most of the paper considers the oracle case in which a reasonably good predictor is available. As discussed in Section~\ref{sec:final}, the very best solutions will likely combine quick model training (AL) with quick annotation (IA). We study the latter and only explore the combination naively by the end of the paper.

The rest of the paper is organized as follows: Section~\ref{sec:related} presents the related work on semi- and self-supervised learning, active learning, and a few key papers related to ours. 
Section~\ref{sec:general} introduces the general binary questioning case,
its relation with information theory, and the solutions given by Huffman encoding and tree search. The proposed method based on rollout is introduced in Section~\ref{sec:practical}, along with the heuristics that make it computationally feasible.
Section~\ref{sec:exp} presents the experimental results on synthetic and real datasets for the oracular and the more realistic HILL case.
Lastly, we revise the limitations of our method, explore its implementation into a tool and conclude our work in Section~\ref{sec:final}.

The contributions of this work are the following:
\begin{enumerate}
    \item We theoretically link IA to Huffmann encoding and explain why it cannot be a feasible solution.
    \item We propose the first intelligent annotation (IA) method for binary classification that reduces annotation cost by $\approx 20\% \text{ to } 88\%$  in the oracle case and by $\approx 70\%$ in the harder from-scratch setting.
\end{enumerate}

\section{Related work}\label{sec:related}

\subsection{Active Learning}
AL methods are commonly evaluated by looking at the \textit{test performance vs. number of annotations} curve \cite{settles2009active-alreview}. Most research explores methods to select  the observations to annotate next to optimally help model training. The methods can be understood through the lenses of Bayesian estimation and information theory \cite{kirsch2022unifying-oxford}. The number of observations to be selected is called \textit{budget} and it is usually big, e.g. 5000 labels \cite{zhan2022comparative-alsurvey}. Batch AL is especially relevant because neural networks are trained using batches and more importantly, because in practice it is more convenient to ask the annotator to annotate many observations at once.

One underlying assumption in many AL works \cite{ash2021gone-al1, NEURIPS2022_f475bdd1-al2} is that the annotation cost of an observation is constant, irrespective of the type of problem, e.g. regression or classification, and irrespective of the complexity of the annotation, e.g. multi-class with 4 vs. 1000 classes. Although the annotation cost of an observation is not constant in most of the cases, it is useful to assume so to keep the focus on the sample efficiency problem.

As will be seen in Section~\ref{sec:final}, AL is an important component of a full solution to the HILL problem. What AL has been missing and this work studies is the annotation advantage one can get by using the predictor being trained. Both AL and IA can be studied simultaneously in the binary classification problem, where the number of annotations equals the number of yes/no answers in the AL case.
In fact, we integrate the most traditional AL methods, namely entropy, least confidence sampling and margin sampling, into our IA solution. Note that in the binary classification case all these are reduced to selecting the samples whose assigned probabilities are closer to $0.5$. We note that even though many other methods exist \cite{zhan2022comparative-alsurvey, kirsch2022unifying-oxford}, simple entropy-based AL is still extremely competitive \cite{beck2021effective-distil}.  

Our work complements active learning in that it can be used to fullfil each one of the batched requests AL yields. However, we deviate from AL with an assumption that is not true in general and is opposite and incompatible with the assumption made in AL: to study intelligent annotation we assume that the annotation cost is given by the number of yes/no answers the annotator needs to provide. For instance, if the annotator is asked ``are these 3 observations positive?'' it could suffice with one ``yes'' to get 3 annotations and the annotation cost of that set of 3 observations would be 1. To answer that question the annotator should look at all annotations at once, and the actual time it takes to get the answer is most likely somewhere between the cost as in AL ($=3$) and the cost as in IA ($=1$).


\subsection{Semi and self-supervised learning}
Semi-supervised learning can improve model training via the use of the unlabeled data\cite{zhu2005semi-semisurvey1, van2020survey-semisurvey2}. One important strategy in semi-supervised learning is self-training or pseudo-labeling. Although we regard these as ways to further improve model training, pseudo-labeling is interesting from the data annotation point of view. For instance, the recently released SAM dataset \cite{kirillov2023segment-sam} is fully composed of automatic annotations. 

Lastly, self-supervised learning \cite{balestriero2023cookbook} can be roughly divided into noisy autoencoding and contrastive learning. Contrastive learning is indirectly supervised \cite{cabannes2023active-grafo} via the definition of the invariance and equivariance conditions the predictor should satisfy. We do not consider data augmentation nor contrastive learning to limit the amount of supervision but we note that prior knowledge is still present for instance in architectural choices or weight-initialization strategies. Of course, if such supervision is available in practice its use is recommended. Similarly, we consider the self-supervised techniques of noisy autoencoding as smart weight-initialization procedures. Semi-supervised learning and self-supervised learning are still very active fields that aim to improve weight initialization and model training. For simplicity, this work ignores most of those advances and builds upon traditional blocks, namely a ResNet18 \cite{he2016deep-resnet}, the Adam optimizer \cite{kingma2014adam} and the binary cross-entropy loss, but a good product should integrate everything.

\subsection{Intelligent annotation}

There are two notable theoretical analysis of this problem. The first \cite{kulkarni1993active-jm1} formalizes the problem in a way that inspires the one presented in Section~\ref{sec:general}.
A key piece of the formalization is the introduction of a \textit{concept} $c$. A way to see it is to let the concept $c$ be the subset of the samples $X$ composed by the samples with positive label. Then there is a one-to-one mapping between $c$ and binary vectors $y\in [0,1]^{|X|}$, where $y$ is the vector of labels $(y_1, \dots, y_{|X|})$ for the full dataset. This is the bridge between the formalization in \cite{kulkarni1993active-jm1} and ours.
Learnability conditions are derived in \cite{kulkarni1993active-jm1} for the fixed distribution case and for the free distribution case. The proof for the fixed distribution case involves a key assumption: the learner wants to minimize the length of the longest codeword rather than the mean codeword length. In other words, \cite{kulkarni1993active-jm1} considers the worst case number of questions. Our work considers the expected number of questions (or mean codeword length) instead. Furthermore, in our framework the minimum number of binary questions is $1$ (the case of automatic annotation with a perfect predictor). 
The second theoretical work \cite{yang2010bayesian-jm2} also deals with (active) learning bounds. The authors correctly point out the equivalence between the problem and coding theory, but choose to consider lossy coding algorithms. The biggest (theoretical) difference between our work and \cite{kulkarni1993active-jm1, yang2010bayesian-jm2} is the change of focus from quick learning to quick annotation and the use of a lossless compression method.

The second class of works are ad-hoc methods such as \cite{xie2022towards, NEURIPS2020_05f971b5-onebit} that introduce yes/no answers as supervision. Notably \cite{NEURIPS2020_05f971b5-onebit} uses binary questioning to reduce the annotation cost of multiclass classification instances. 
Similarly, \cite{xie2022towards} aims to reduce the cost of annotating an observation but for the image segmentation case. 

\subsection{Dynamic Programming}
Dynamic Programming (DP) is an optimization algorithm widely used to solve optimal control / reinforcement learning problems. The closest to our work is the recent Wordle resolution \cite{bhambri2022reinforcement-wordle} based on the rollout method popularized by Bertsekas \cite{bertsekas2012dynamic-popularized2, bertsekas2022rollout-popularized1}. Rollout is a reinforcement-learning method that yields an improved policy based on an initial policy \cite{bertsekas2021rolloutbook}. For some control problems there is an heuristic policy based on a proxy cost function that serves as initial policy.
The main idea of rollout is that, for these cases, one can use lookahead minimization of the same proxy cost function by going down the tree of actions/states and obtain a new policy that is guaranteed to be an improvement over the original heuristic.

\section{General setting}\label{sec:general}
Given a dataset of size $N$ we (re-)define the set of observations $X=(x_1, \dots, x_N)$ and the associated ground truth labeling $y=(y_1, \dots, y_N) \in [0,1]^N$. As shown in \cite{kulkarni1993active-jm1} a general binary question is of the sort ``\textit{is $y \in q$?}'' where $q$ is a subset of the set of possible labelings, i.e. $q\subset [0,1]^N$. In particular the annotation of a single sample can be phrased as ``\textit{is $y \in q_{y_i=1}$?}'' where $q_{y_i=1}=\{z \in [0,1]^N | z_i=1\}$. The problem of annotating the full dataset with the minimum number of questions is exactly the problem of encoding $y$ with the minimum expected length. The sequence of Q\&A is given by the corresponding decoding process. The source coding theorem establishes that the minimum expected length achievable by a lossless compressor is given by $H(Y)$ \cite{shannon2001mathematical}, where $Y$ is the random variable associated to the labeling $y$. 
This is a lower bound on the number of questions $Q$ that we can compute if we know the distribution $P(Y=y)$. For our purposes we assume that (i) $P(Y=y|X)$ is known and that (ii) the entries of $y$ are independent, i.e. $P(Y=y|X) = \prod_i P(Y_i=y_i|x_i)$. These assumptions fit the familiar case where a predictor $f_\theta$ can estimate the probabilities $f_\theta(x_i) = \hat{P}(Y_i=1|x_i)$. We will now use $f_\theta$ in place of $P$ but this is only valid in the oracle case. The entropy of the distribution is 
\vspace{-4mm}
\begin{multline}
    H(Y)= -\sum_{y\in Y} P(Y=y)\log(P(Y=y)) \\
        =-\sum_{1\leq i \leq N}\left[ f(x_i)\log(f(x_i)) + (1-f(x_i))\log(1-f(x_i))\right].
\end{multline}
\vspace{-4mm}
\subsection{Huffman encoding}
Huffman encoding \cite{huffman1952method} is a lossless encoding method that is optimal for the case of single-symbol encoding. Optimally encoding a sequence of symbols is a different problem often solved with arithmetic codes. Huffman encoding achieves the entropy bound when the probabilities assigned to every symbol are dyadic, i.e. of the form $1/2^k$. The algorithm works as follows: (i) set the initial list of nodes without a parent to be all symbols. The symbols in our case are all possible dataset labelings, i.e. the $2^N$ binary vectors of length $N$. Then (ii) the two nodes without a parent with the smallest probability are assigned a new node as parent whose probability is the sum of the probabilities of the children. Step (ii) is repeated until only one node is left without parents. This last node is an ancestor for all symbols. Once the tree is built, each node represents a state given by all labelings that have not been discarded, which are the descendant symbols of that node. 

The decoding process is conducted as follows: starting from the root node we ask the question ``\textit{is $y \in q$?}'' where $q$ is the right branch coming out of the node. Repeating this step takes us down the tree until only one symbol is left. We are guaranteed that, given the probabilities $P(Y=y)$, we will arrive at the ground truth labeling $y$ in the minimum expected number of questions Q (this is proven via an exchangeability argument and induction) \cite{berkeley}. Furthermore, one can prove that $H(Y)\leq Q_{\text{huffman}} \leq H(Y) + 1$ \cite{mit}. The computational complexity for building the tree is $m\log m$ where $m$ is the number of symbols in the alphabet. In our case $m=2^N$. This is exponential complexity and becomes unmanageable for values of $N$ as small as $N=20$. 

\begin{figure}
    \centering
    \includegraphics[width=0.6\textwidth]{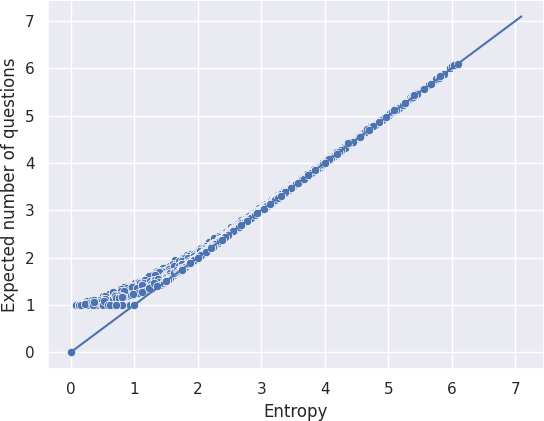}
    \caption{Computed expected cost on Huffman tree vs. entropy for the synthetic problem (a).}
    \label{fig:huffman}
\end{figure}

Given the Huffman tree and the probabilities we can compute the expected number of questions from a given node. In Figure~\ref{fig:huffman} we show the comparison between this exact computation and the entropy, finding that indeed the expected cost on the Huffman tree gets very close to the entropy bound. Lastly, note that because Huffman encoding falls into the general questioning case, the questions are of the sort \textit{is the correct labeling of the whole dataset $y$ among those labelings in $q$?}. The user experience when answering this question is not great, and this will be fixed with an heuristic in Section~\ref{sec:practical}. 

\subsection{Exhaustive tree search and DP}
If we take the dyadic case where Huffman encoding reaches the entropy bound we have that, simultaneously, there is a policy or questioning strategy (Huffman) for which $E[Q]=H(Y)$ and also we have that $E[Q]\geq H(Y)$. It follows that the optimal value or cost function, i.e. the evaluator function that returns the expected cost incurred by the optimal policy from a given state, is the entropy $H(Y)$. Given such an optimal value function the derivation of the optimal policy is simple: we choose the action (a subset $q$ of binary vectors) that minimizes the expected entropy on the next state. In other words, given a state $s_t$ we consider all possible questions $q$ and choose $\text{argmin}_q J^\star(s_t, q)=\text{argmin}_q J^\star({s_{t+1}}_1|a=1)P(a=1) + J^\star({s_{t+1}}_0|a=0)P(a=0)$ where $J^\star = H(s)$. Note that to compute $H(s)$ we need to update the probabilities $P(Y=y)$ using Bayes rule to restrict them to $s$, which simply accounts for a renormalization. This method gives us a procedure that is optimal and is top-down (greedy) instead of bottom-up (Huffman). The computational complexity is given by that of the entropy computation times the number of possible questions. The number of possible questions in the first step is the cardinality of the set of all possible subsets of binary vectors of length $N$, this is, the cardinality of the power set $2^{([0,1]^N)}=2^{2^N}$. This exponential-exponential computational cost is way higher than the exponential of Huffman and also unfeasible.

\section{Practical setting}\label{sec:practical}   
The optimal solutions described above have two issues that make them not useful in practice. The first is the computational complexity, which is (at least) exponential on the dataset size. The second is the type of binary questions, which are general and over all the dataset, and make the user experience of any product unacceptably bad. Our proposed IA method solves these two issues.

The proposed method is based on two key observations, namely that lookeahead minimization guarantees improvement over the base heuristic and that the branching factor of the tree (originally $2^{2^N}$ should be drastically reduced. Another quantity of interest that will be used in what follows is the size $|s|$ of the state $s$, which represents the number of labelings that were not yet discarded. 

\subsection{Reducing the computational cost}
To reduce the branching factor of $2^{2^N}$ to something reasonable we use a handful of heuristics, each one taking us further away from optimality but closer to an efficient method. We note that all possible questions can be counted considering the cardinality of the questions $n=|q|$ as $\sum_{1 \leq n \leq 2^N} \binom{2^N}{n}= 2^{2^N}$. These are:
\begin{enumerate}
    \item \textbf{Early stopping}: We will assume that the cost of the best question of size $n$ as a function of $n$ has a global minimum. We look for questions starting from $n=1$ and increase $n$ but stop once we find that the cost of the best question of size $n$ is higher than the cost of the best question of size $n-1$. This reduces the branching factor to $\sum_{1 \leq n \leq E} \binom{2^N}{n}$ where $E \ll N$.
    \item \textbf{Select most likely}: Not true in general, we consider that the best question of size $n$ is given by the $n$ most likely labelings of those that remain available in $s$. This criterion is optimal when the cost function is $|s|$, but it is not optimal in general for the case of the entropy as a cost function. This heuristic reduces the branching factor to $\sum_{1 \leq n \leq E} 1 = E$, if the $2^N$ possible labelings are sorted and to $2^N \log(2^N) E$ if not.
\end{enumerate}
We now introduce two more heuristics that are not only motivated by the computational complexity problem but also by user experience considerations and algorithmic simplicity:
\begin{enumerate}
\setcounter{enumi}{2}
    \item \textbf{Question as single guess}: This heuristic targets user experience: the space of questions is now given by the (most likely) guesses over $n$ observations (abusing notation of $n$). This means that the method will choose some set of $n$ observations, pseudo-label them, and ask the annotator if the pseudo-labeling is correct. We maintain the early stopping and most likely heuristics that now cause our search to be of complexity $N \log(N) E$, caused by sorting the probabilities for each observation and evaluating guess sizes $n$ from $1$ up to $E$ (we early stop when the cost function increases). 
    \item \textbf{Don't give up}: Until the introduction of the single guess heuristic, the incorrect guesses were simple to handle: we simply removed them from the state. In contrast, when using guesses the state can be summarized as the current unlabeled indices and the incorrect guesses made so far. Updating the state given a new correct or incorrect guess is possible but the simplification of a state (without redundancy) is somewhat contrived. Although we successfully implemented general state updates we prefer to introduce the \textit{don't give up} heuristic for simplicity. The \textit{don't give up} heuristic establishes that when an incorrect guess of size $n$ has been made, the next guess should be the incorrect guess with the most uncertain prediction removed. When the second guess is correct, we can deduce the labels for all the indices of the first guess. In contrast, when the second guess is incorrect, we apply the heuristic again. This reduces the branching factor to $N\log(N)$ when an incorrect guess has been made. Note that the sorting can be made only once if the predictor $f$ stays the same.
\end{enumerate}

\paragraph{Lookahead minimization}
The heuristics introduced above can successfully reduce the number of possible questions to be considered at each step. However, choosing the one that minimizes the expected entropy or the expected size on the next state is not an optimal method anymore. Optimality was lost when reducing the question space. However, using lookahead minimization, i.e. simulating a few steps ahead and only then evaluating the states using the chosen proxy cost function, is guaranteed to improve over the heuristic generated by the proxy cost function \cite{bertsekas2022rollout-popularized1}.
In the limit of maximum lookahead the terminal states are reached and the exact expected number of questions is computed, thereby achieving the optimal method respecting the heuristics above.

\paragraph{Summary and parameters}
In a nutshell, our method (IA) is lookahead minimization over the tree of state and actions. Some parameters control the algorithm, in italics in this section. The tree search involves selecting a node, expanding the selected node, updating the parents of the expanded node, and after a number of expansions taking the action that has the best expected cost. When expanding a node, guesses with different sizes $n$ are considered as actions, with a special treatment for $n=1$. In fact, for $n=1$, we allow to choose between \textit{random sampling or entropy sampling}, but any active learning algorithm could be used. To allow for more control on the branching factor and the user experience a parameter that sets the \textit{maximum allowed guess size $n$} is added. Another parameter of our method is the \textit{maximum node expansions} to be made at each step that roughly controls the computational cost of the tree search.

The selection of the node to be expanded is done according to the priority, where the priority is the product of the probabilities of the edges. Even though we always use a greedy approach for action selection, to set priorities we consider the probability of an action to be the result of the softmax over the cost of the different actions with some \textit{temperature}. Lastly, a \textit{proxy cost function} is used, and we allow the choice between the \textit{entropy} $H(s)$ and the log state size $\log_2(|s|)$ (roughly the number of points left to annotate). Both functions are adapted for the case in which an incorrect guess is part of the state by using conditional entropy and counting concepts. 

After each node expansion the parents are updated, which means that we propagate the new estimate of the state cost up the tree and continue expanding nodes if needed. Of course, if the probabilities given by $f$ do not change, after each step the relevant branch of the tree is kept in memory, thereby allowing the \textit{max expansions} to further expand the tree. We refer the reader to the supplementary material for full code and algorithm description.

\paragraph{Fitting to incorrect examples}
In the online learning case where the predictor $f$ is continuously being learned, the predictor can and should be updated if an incorrect guess has been made. The simple method used to do this is now described. Consider $\bar{x} = (x_1, \dots, x_n) \in X$ without loss of generality when selecting the indices and the associated predicted labeling $\hat{y}=(\hat{y}_1, \dots, \hat{y}_n)$. The estimated probability of $\hat{y}$ being correct is given by $\hat{P}(\hat{y} \text{ ok}) = f_J(\hat{y} \text{ ok}) = \prod_{1\leq i \leq n} f(x_i)^{\hat{y}_i}(1-f(x_i))^{1-\hat{y}_i}$, where $f_J$ assigns a probability to each answer. It follows that given an incorrect guess the binary cross entropy between $f(\hat{y}\text{ ok})$ and the actual answer $0$ can be used as a loss that can be minimized via gradient descent. This loss is added to the loss computed over the labeled examples for each epoch and in this way the knowledge of the guess being incorrect is integrated into $f$.

\section{Experiments}\label{sec:exp}
Here we summarize our main results. The Appendix includes full numerical results, experimental details and extra figures.


\paragraph{Synthetic}

\begin{figure} 
\centering 
\subfloat[Two Gaussians. The predictor $f=P(Y|X)$ follows Bayes' rule.]{
\includegraphics[width=0.2867\textwidth]{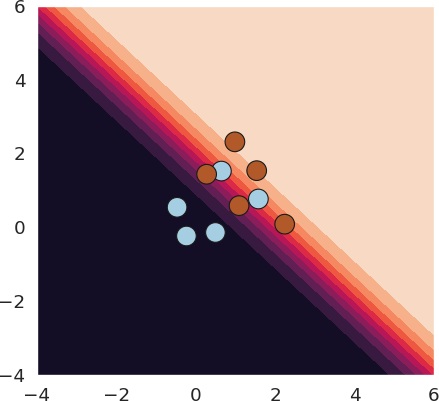}
\label{fig:toy1} 
}
\hspace{1em}
\subfloat[Two Gaussians. The predictor $f(X)=P(Y|X)$ is independent of $P(X)$.]{
\includegraphics[width=0.2670\textwidth]{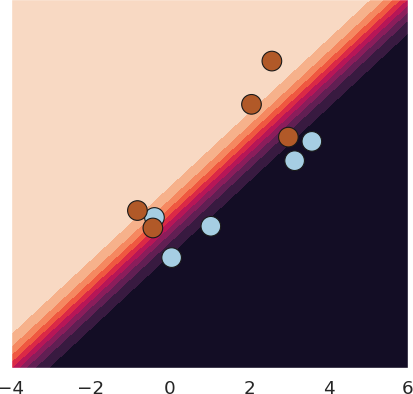} 
\label{fig:toy2} 
}
\hspace{1em}
\subfloat[Data following \cite{guyon2006design}. The predictor $f$ is a logistic regressor trained on the data and $f(X)\neq P(Y|X)$.]{
\includegraphics[width=0.337\textwidth]{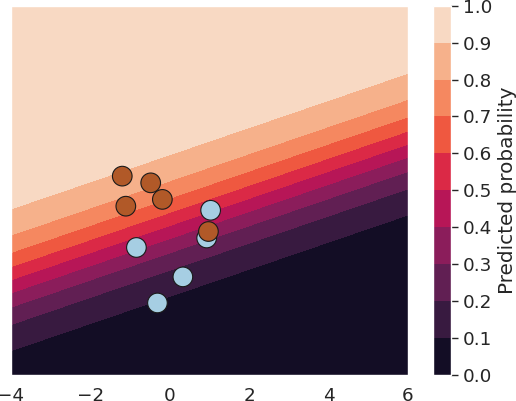} 
\label{fig:toy3} 
}
\caption{Example sample of three two-dimensional synthetic binary classification problems. The background color corresponds to the probabilities predicted by $f$. The color of the points correspond to the ground truth labeling.} 
\label{fig:toys} 
\end{figure} 

\begin{figure} 
\centering 
\subfloat[]{
\includegraphics[width=0.33\textwidth]{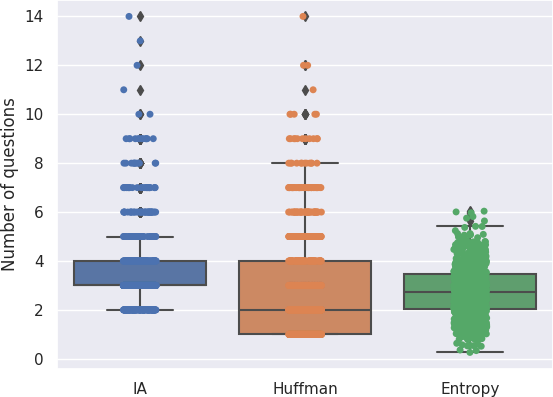}
}
\subfloat[]{
\includegraphics[width=0.33\textwidth]{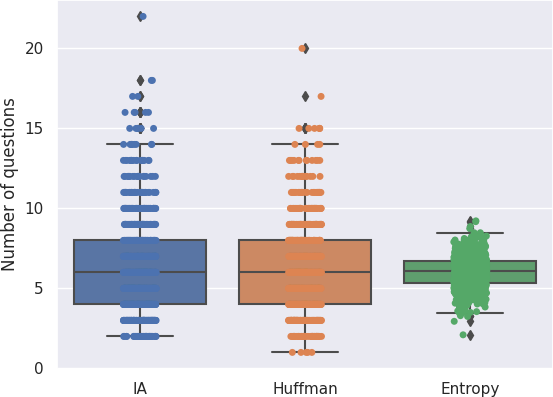} 
}
\subfloat[]{
\includegraphics[width=0.33\textwidth]{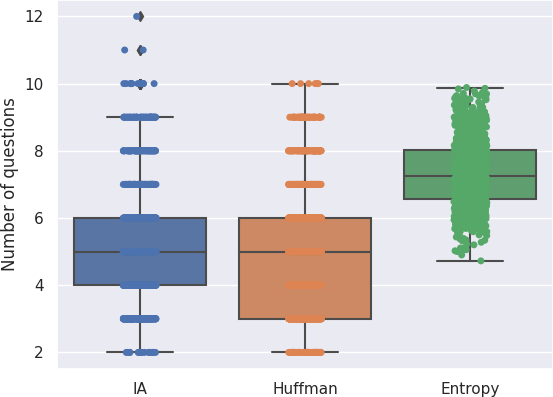} 
}
\caption{Number of questions for the synthetic problems (ref. Figure~\ref{fig:toys}) for each method (IA, Huffman) and the entropy of the data as estimated by the predictor.} 
\label{fig:boxwhisker1} 
\end{figure}

We generate three two-dimensional toy problem variations with dataset size $N=10$ for $1000$ different seeds. The variations are depicted in Figure~\ref{fig:toys}. In Figure~\ref{fig:boxwhisker1} one observes that the proposed method is very close to the optimal given by Huffman ((a),(b) and (c)), that is in turn close to the entropy lower bound ((a) and (b)). For the last problem (c) we observe that the entropy computed by the predictor that is fit to the data but does not capture the underlying distribution $P(Y|X)$ does not act as a lower bound. This means that pretrained predictors can incorrectly estimate the complexity of the data they were fitted to.


\begin{figure} 
\centering 
\subfloat[CIFAR10 \cite{krizhevsky2009learning-cifar10}]{
\includegraphics[width=0.5\textwidth]{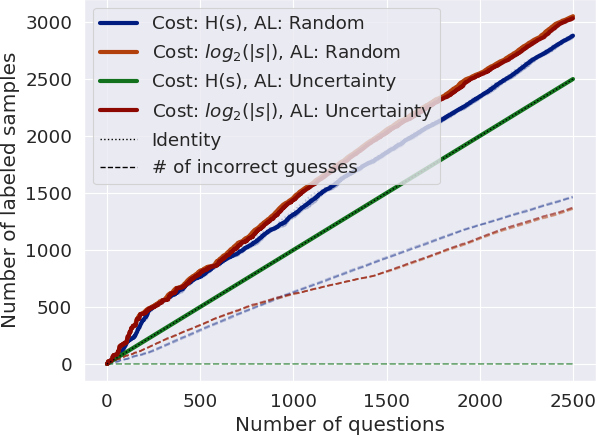}
\label{fig:cifar} 
}
\subfloat[SVHN \cite{netzer2011reading-svhn}]{
\includegraphics[width=0.5\textwidth]{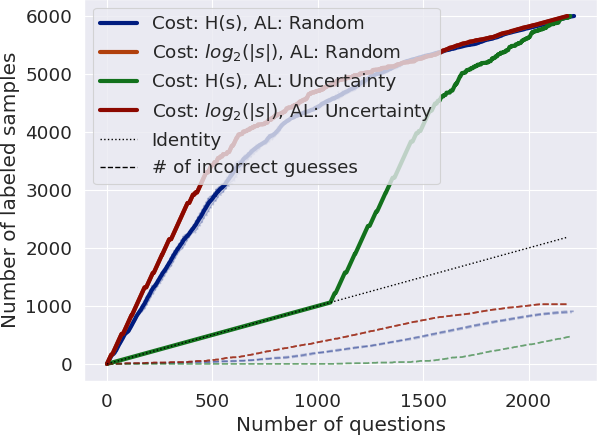}
\label{fig:svhn} 
}
\caption{Number of annotations vs. number of questions for four IA variants on the first 6000 examples of each dataset and leveraging a fixed pretrained model.} 
\label{fig:pretrained} 
\end{figure}

\paragraph{Real with pretraining}
The IA method is evaluated over the first 6000 samples of the CIFAR10 \cite{krizhevsky2009learning-cifar10} and SVHN \cite{netzer2011reading-svhn} datasets, where all labels with indices greater than 4 are considered positive. These datasets are too hard to run from scratch and a fixed predictor trained on different samples of the training set is provided to the IA. The oracles have accuracies of $0.67$ and $0.93$ respectively. Figure~\ref{fig:pretrained} shows the number of annotated samples vs. the number of binary questions posed. The maximum annotation gain (number of answers required to annotate $2500$ points) for CIFAR10 is $20\%$, which is expected considering that the predictor is not very good. In contrast, for SVHN all methods manage to annotate the full $6000$ samples with less than $2500$ questions, showing how effective the method can be when the predictor is good. We note that the number of incorrect guesses is permanently increasing in most cases, indicating that the size of the guesses is larger than one. Furthermore, when using the entropy of the state $H(s)$ as a proxy cost function and offering the most uncertain sample as the potential guess with unit size, the IA method sticks to point-by-point annotation for more than $1000$ examples. This variant then switches to annotating the most obvious datapoints at a high annotation rate.

\paragraph{Real from scratch}
Then we turn our attention to the more realistic case of intelligent annotation when a trained predictor is not available. In this case the predictor is trained from scratch with the available annotations so far. This is the case (dubbed ALIA) that combines the related tasks of active learning (quickly training a predictor) and intelligent annotation (using that predictor to annotate more efficiently) and yields both annotated data and a trained predictor. For this experiment we run the ALIA over MNIST \cite{lecun1998gradient-mnist} and Fashion MNIST \cite{xiao2017/online-fmnist}. An annotation gain of $70\%$ is observed for both datasets. As expected, after most easy samples are annotated, the annotation slows down. This change is very abrupt for MNIST and we hypothetize that this is due to an uncalibrated predictor.

\begin{figure} 
\centering 
\subfloat[MNIST \cite{lecun1998gradient-mnist}]{
\includegraphics[width=0.5\textwidth]{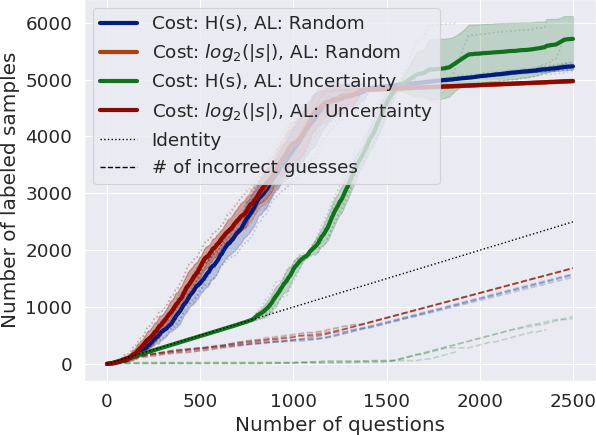}
\label{fig:sub14} 
}
\subfloat[FMNIST \cite{xiao2017/online-fmnist}]{
\includegraphics[width=0.5\textwidth]{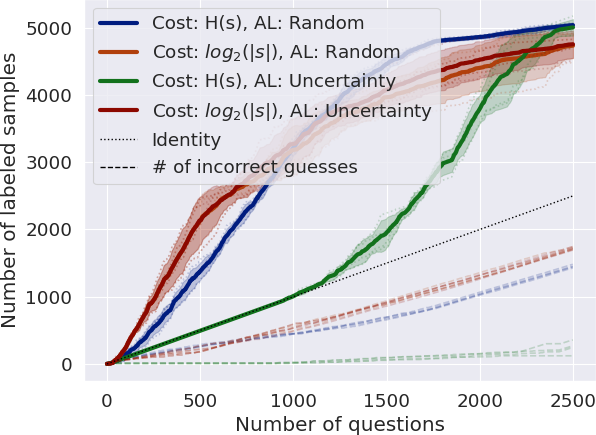}
\label{fig:sub24} 
}
\caption{Number of annotations vs. number of questions for four ALIA variants on the first 6000 examples of each dataset starting from a randomly-initialized predictor.} 
\label{fig:whole5} 
\end{figure} 

\section{Product, Limitations, Conclusion}\label{sec:final}
Our method naturally suggests a software for the annotation of binary classification. The annotator is presented with a classification of at most \texttt{max\_n} samples and they answer \textit{correct} or \textit{incorrect}. The classification is the pseudo-labeling of a predictor $f$ that is continuously trained in the background using all the annotated data so far and checkpoints are frequently saved. The last checkpoint is loaded by the tree search that uses its predicted probabilities to select the number $n$ of samples presented to the annotator as described in Section~\ref{sec:practical}.

Our work is arguably the first attempt to solve the intelligent annotation problem, with some limitations. One of them is that, similarly to most AL work, the present work mostly ignores the interplay between quick annotation and quick learning, where \textit{quick} refers to \textit{the number of binary answers}. It is natural to expect that for both problems a sweet spot will be found in between, i.e. the best solution will involve quickly annotating data that makes the predictor training quicker. Furthermore, all our experiments have dealt with balanced datasets, while in many real scenarios only imbalanced data is available. Another limitation is that we have not considered in-depth the impact on our method of imperfect predictors, in particular uncalibrated predictors. Even though the method works in practice, the whole tree search is dependent on probabilities that might not correspond to true frequencies. 
Lastly, even though the computational complexity has been greatly reduced, the method becomes slower for large $N$ (freezing the predictor and reducing the number of node expansions makes the method fast in practice). 

Motivated by the need of tools for quick annotation, this paper studies how a predictor can be leveraged to quickly annotate data and measures quickness as the number of yes/no answers provided by the user. We show that the optimal binary questioning strategy given a perfect predictor is found in Huffman encoding, but this strategy is too computationally expensive to be applied in practice. We manage to develop a practical intelligent annotation (IA) solution by introducing heuristics and relying on the dynamic programming principle of lookahead minimization. We show that our strategy works comparably to the optimal one on toy problems and that it provides significant annotation efficiency gains over more realistic data.

\printbibliography

\newpage

\appendix

\section{Code}\label{sec:code}
Code is available at \url{https://anonymous.4open.science/r/optimalbinaryhill/}.

\section{Proposed method}
Even though the exact code is provided, here we provide Python-styled pseudo-code for the proposed method to ease understanding. The following function is the one that chooses the best guess, question or action from a state node. 

\begin{python}
def tree_search(root_node):
    print("Expanding tree...")
    for i in range(self.max_expansions):
        node = select_node(root_node)  # highest priority leaf node
        if node == "all nodes expanded":
            break
        # get the predictions relevant to the current node
        expand_node(
            node,
            max_n,
            node_predictions,
            al_method,
            approx_cost_function
        )
        update_parents(node)  # here

    best_action_node = get_best_action(root_node)
    return best_action_node 
\end{python}

Here \verb|root_node| is a state node, described by a list of indices corresponding to the unlabeled datapoints and optionally an incorrect guess. As described in the manuscript, \verb|max_n|, \verb|al_method| and \verb|approx_cost_function| are parameters describing the maximum number of elements in a guess, the way to choose the datapoint when the number of elements in a guess is $1$, and the choice between the entropy of a state and the log of the number of possible labelings in a state, respectively.  

The algorithm is fully described by the definition of the functions \verb|select_node|, \verb|expand_node|, \verb|update_parents| and \verb|get_best_action|. These are fully described in Section~\ref{sec:code} but a simple description follows:
\begin{itemize}
    \item \verb|select_node|: This function explores the tree and sets a priority for each node respecting the following rules: (i) if the node is the root node, the priority is $1$, (ii) the priority of an action node is the priority of its parent state node times the softmax with temperature $10$ of the sibling action costs at the particular action index and (iii) the priority of a state node is the priority of its parent action node times the probability of the transition to the state. In short, we consider priorities depending on probabilities and action values. Once the priorities are set the leaves under a \verb|MAX_DEPTH=20| are listed and the one with highest priority from the ones that are not terminal is selected.
    \item \verb|expand_node|: The state node expansion involves creating action nodes and assigning an initial estimation of their value. The actions are guesses of lengths represented by $n$. Firstly, $n=1$ and the guess is selected according to the \verb|al_method| selected (random sampling or highest uncertainty). For this action, the computation of the next state is trivial as it is simply adding the label of the guessed observation to the list of current labels. If an incorrect guess was present it is removed providing one annotation for free if the guess was correct. This free annotation comes from the fact that the last guess, given that an incorrect guess was present, is precisely the incorrect guess with one less observation. For the case of a guess $n=1$ it means that the incorrect previous guess had $n=2$.
    Then, if the last guess is correct it means that the extra label from the incorrect previous guess was wrongly guessed and we can deduce the actual label is the opposite.     
    
    The second step is to increment $n$ and select the subset of the $n$ \verb|node_predictions| with most certainty. The probability of this guess being correct can be computed via multiplying the probabilities of correctness of each single label where the probabilities are adjusted if an incorrect guess was present (these computations are derived using Bayes' rule). These probabilities are associated to two possible next states which correspond to (if correct) adding labels and removing the incorrect previous guess if any and (if incorrect) replacing the incorrect previous guess with the new incorrect guess.
    
    After a new action node is added then its value is computed as a probability-weighted average of the values of its children states. These values are computed using the \verb|approx_cost_function|, that can be the entropy or the log of the number of possible labelings. 
    We repeat the second step of incrementing $n$ until the cost of the new action node is greater than the cost of the previous action node. The cost of the action with minimum cost is assigned to the state node being expanded.
    
    \item \verb|update_parents|: When a state node is expanded its associated cost is updated. This function updates the cost of the parent action node, if any. Furthermore, if this update changes the optimal action then we continue expanding the grandparent state node (if needed). If it does not change the optimal action but it changes the grandparent state node cost, i.e. if the parent node corresponds to the optimal action, the same update process is repeated from the grandparent state node.

    \item \verb|get_best_action|: This function simply looks at the children action nodes of the root state node and chooses the one with the least cost.
\end{itemize}

The content above explains how to choose the best action given a state. The complete loop involves finding the best action, getting the answer for that question, updating the state, and optionally retraining the predictor. For the oracular case there is no predictor retraining. For the HILL or ALIA case there is retraining ideally after each annotation step. To keep the computational cost low we retrain after every state update if we have made less than $100$ questions, else if we have made less than $200$ questions we retrain every two state updates, and so on.
Full experimental parameters are given in Section~\ref{sec:details}. 

Lastly, we note that when \verb|max_n| is greater than $8$ and depending on the distribution of the optimal $n$ we might get a more efficient method by looking for $n$ with an exponential search.

\section{Experimental details for synthetic datasets}\label{sec:details}

\subsection{Datasets}
The three synthetic datasets generated are two-dimensional. 
\begin{enumerate}[(a)]
    \item \textbf{Two blobs}: 5 points are sampled from each one of two gaussians. Both have identity covariance matrices and they are cenetered at $(0,0)$ and $(2,2)$ respectively. The likelihood is given by $P(X=x|y)= \frac {1} {2\pi} \exp{\left(-\frac {1}{2} ((x_1-c_y)^2+(x_2-c_y)^2)\right)}$ where $c_y$ is the center of one of the gaussians (we set $c_0=0$). Then the probability of an observation $x$ being positive is
    \begin{multline}
    P(y=1|x) = \frac{P(x|y=1)P(y=1)} {P(x, y=1) + P(x, y=0)} \\
    = \frac{P(x|y=1)P(y=1)}{ P(x|y=1)P(y=1) + P(x|y=0)P(y=0)} \\
    = \frac{P(x|y=1)}{ P(x|y=1) + P(x|y=0)} 
    \end{multline}
    where $P(y=1)=P(y=0)$ because we create a balanced dataset. The function $P(y=1|x)$ is used as the predictor.
    \item \textbf{Two blobs with linear predictor}: To show that our method does not rely on clustering properties (unlike some semi-supervised methods \cite{zhu2005semi-semisurvey1}) we again create datapoints from two blobs but now assign the labels according to a probability given by a pre-defined predictor. In this case $P(X)$ corresponds to a balanced gaussian mixture with centers $(0,0)$ and $(3,3)$ and covarianecs given by $Id / 2$. The $P(y=1|x)$ is given by 
    \begin{equation}
        P(y=1|x) = \frac{1}{1+\exp{\left(-\frac{x_2 - x_1}{2} / 0.3\right)}}
    \end{equation}
    and each label $y_i$ is set to $1$ with probability $P(y_i=1|x_i)$.
    \item The last case is generating points using the \verb|make_classification| function of \texttt{scikit-learn} that follows \cite{guyon2006design}. Our parameters are \verb|n_features=2|, \verb|n_classes=2|, \verb|n_informative=1|, \verb|n_redundant=0|, \verb|n_repeated=0|, \verb|n_clusters_per_class=1|, \verb|flip_y=0.2|. The result is that the blobs have different covariance and there is some flip noise inserted. Our predictor in this case is different. Instead of relying on the true label probabilities we train a predictor (for each sample of $10$ points). This predictor is the \verb|LogisticRegression| classifier from \texttt{scikit-learn}. This is a linear classifier with a sigmoid function and L2 regularization, trained with the lbfgs solver that uses gradient descent and quasi-Newton method. Because it is a learned predictor the $P(y=1|x) \neq f(x)$ and that causes the qualitatively different results seen in Figure~\ref{fig:boxwhisker1}.
\end{enumerate}

\subsection{Parameters}
IA parameters for our experiments on synthetic datasets include: 
\begin{python}
TEMPERATURE = 10
MAX_DEPTH = 20
al_method = "uncertainty"    
max_n = 8
max_expansions = 8
cost_fn = "entropy"
reduce_certainty_factor=0.01
reset_tree=False
\end{python}

The Huffman questioning method has no parameters.

\subsection{Experiment details}
We run the same experiment for $1000$ different seeds $(0, \dots, 999)$. 
There is no initial labeling provided as the predictors are fixed for each seed. Note that because the tree is not reset it (most likely) grows after every question.

\section{Experimental details for real datasets}\label{sec:detailsreal}
\subsection{Datasets}
The datasets used for the experiments are CIFAR10 \cite{krizhevsky2009learning-cifar10}, MNIST \cite{lecun1998gradient-mnist}, FMNIST \cite{xiao2017/online-fmnist}, SVHN \cite{netzer2011reading-svhn}. The first three are balanced, while SVHN has some bias towards smaller digits. 
The datasets are binarized by setting a positive label if the original label is greater than $4$. This binarization mantains the (un)balanced character of the datasets. 
The datasets are downloaded using \verb|torchvision.datasets| API. The native \verb|Dataset| class is substituted by a custom class that stores the dataset in memory for fast dataloading.
For our experiments we use the first $6000$ examples from the training set of each dataset.

The color datasets are normalized using ImageNet mean and standard deviation and the single-channel datasets are normalized using MNIST mean and standard deviation.

\subsection{Model parameters}
Below are the model parameters of our IA solution. All these were created to provide flexibility to the user and set to (roughly) arbitrary values. Preliminary experiments suggested that the temperature set at the order of 10 provided a good balance between breath and depth expansion of the tree. The maximum depth was set to $20$ because planning $20$ steps ahead was already considered too much. For the case in which the tree is not reset, i.e. when the predictions do not change, the tree grows at each iteration and the maximum depth parameter is a way to avoid unlimited growth and potentially high memory consumption. We do not expect this limit to be reached. The last parameter is the factor that reduces the certainty of the predictor via a weighted average with $0.5$. This parameter was introduced because the predictions can be sometimes too confident. A factor of $0.01$ should not significantly modify the behavior of the model, while a parameter of $0.05$ is already enough to make the length cost function somewhat less aggressive. We did not run ablation studies on this factor for any of the two losses.

\begin{python}
TEMPERATURE = 10
MAX_DEPTH = 20
reduce_certainty_factor=0.05 if cost_fn=='length' else 0.01
max_n = 8
max_expansions = 8
\end{python}

The \verb|max_n| parameter and the \verb|max_expansions| parameters were set to $8$ to maintain a good tradeoff between speed and performance. Note that the number of expansions accumulate when the tree is not reset and the branch selected by the question and outcome contains at least one node expansion.  

\subsection{Architecture and pretrained predictor}
The architecture used is a ResNet18 \cite{he2016deep-resnet} from \verb|torchvision.models| with \verb|num_classes=1|. The number of input channels is set to $3$ and when the input image is single-channel (e.g. MNIST) the single-channel is replicated two more times. The maxpooling layer from the ResNet18 is deleted to account for the image size difference with respect to ImageNet. The probabilities are obtained by applying a sigmoid function to the output of the ResNet. We tried compiling and running the predictor using \verb|torch.compile| but it resulted in slower runtimes. 

For CIFAR10 and SVHN we pretrain the predictor on all the training datapoints but the first $6000$. The pretraining uses a batch size of $6000$ and Adam optimizer with initial learning rate set to $0.01$. The pretraining comprises $24$ epochs. 

\subsection{Experiment and training}
Each experiment was run for four seeds (0, 1, 2, 3) on GeForce, V100 or Titan Nvidia GPUs as available. The runtimes vary for each experiment but it is always less than $30$ minutes. The speed of the tree search is below $0.5$ seconds per search. 

A first annotated example, selected randomly, is provided. This is one source of stochasticity, albeit minor.
For the case with fixed predictions the deviations between runs is negligible and the error intervals can barely be noticed (e.g. see Figure~\ref{fig:cifar}). We can thus conclude that the stochasticity coming from what is the first annotated example is very low. Most of the stochasticity visualized for the experiments over MNIST and FMNIST comes from the initialization of the predictor weights. 

For these later cases of HILL or ALIA, we retrain the predictor after each question for the first $100$ examples, every two questions for the next $100$ examples, and so on. Each time the predictor is retrained starting from the latest set of weights. The predictor is retrained using the current labeled examples and the current incorrect example if any for $4$ epochs. The optimizer is Adam with initial learning rate $0.01$ and it is reinitialized at each new retraining. The fitting to the incorrect example is done as described in Section~\ref{sec:practical}. This is done without any particular weighting. The incorrect example counts to the loss the same as any labeled sample.
We avoid using batches and thus our gradient descent is non-stochastic. For all experiments, the maximum number of questions considered was $2500$.

\section{Questioning as encoding}
The manuscript reads: ``\textit{The problem of annotating the full dataset with the minimum number of questions is exactly the problem of encoding $y$ with the minimum expected length.}''. However, this might not be obvious for the reader. The problem as it was formulated involves, with some knowledge of the probability $P(Y)$ where $Y$ is the random variable representing the labeling of the full dataset, to find through binary questions what specific value $y$ the random variable takes for our realization. This is, we want to recover a full dataset labeling $y$ with the least number of binary questions. The encoding problem in communication theory tackled by Huffman involves assigning to each possible symbol $y$ a binary code. This binary code is, on average over many samples of symbols, the one that has the least number of bits. The Huffman decoding process can be seen as a sequential question game where we ask ``\textit{is the next bit of the symbol representation a $1$?}''. The answer of each question is a bit that is communicated to the receiver. With the least number of bits (on average) the Huffman method can reconstruct the symbol $y$. Another way of phrasing the question is ``\textit{is the symbol being transmitted on the branch to the right of the current node in the Huffman tree?}'', which is one case of a general binary question (``\textit{is $y \in q$?}''). The equivalence between annotation and communication theory can be drawn also when trying to find the minimum possible expected number of questions needed to annotate a dataset provided with the probabilities of the labels. This is given by the entropy and the reasoning is similar. The key to the reasoning is considering not each individual label $y_i$ but the full labeling $y$ as a symbol.

\begin{table}
\caption{Results over synthetic datasets. All entries are mean values over the quantities in the column titles. $Q$ represents the number of questions and $H$ represents the entropy for a given sample of $N=10$ synthetic datapoints.}
\label{tab:synthetic}
\centering
\begin{tabular}{@{}lccc|ccc|ccc@{}}
\toprule
\cmidrule(l){2-10}
        & \multicolumn{3}{c|}{Synthetic (a)}                   & \multicolumn{3}{c|}{Synthetic (b)}                   & \multicolumn{3}{c}{Synthetic (c)}                    \\ \cmidrule(l){2-10} 
        & $\overline{Q}$ & $\overline{Q-H}$ & $\overline{Q/H}$ & $\overline{Q}$ & $\overline{Q-H}$ & $\overline{Q/H}$ & $\overline{Q}$ & $\overline{Q-H}$ & $\overline{Q/H}$ \\ \midrule
Entropy & 2.77           &        -         &        -         & 6.03           &            -     &         -        & 7.33           &         -        &         -        \\
Huffman & 2.80           & 0.03             & 1.05             & 6.11           & 0.08             & 1.01             & 5.08           & -2.25            & 0.67             \\
IA      & 3.46           & 0.69             & 1.42             & 6.31           & 0.28             & 1.05             & 5.20           & -2.13            & 0.69             \\ \bottomrule
\end{tabular}
\end{table}

\begin{table}
\caption{Performance of methods over CIFAR10 dataset using a pretrained predictor.}
\label{tab:cifar10}
\centering
\begin{tabular}{@{}llrrrrr@{}}
\toprule
\cmidrule(l){3-7}
            &               & \multicolumn{5}{c}{CIFAR10}                                                                                                                                                                                                                            \\ \midrule
AL method   & Cost function & \multicolumn{1}{l}{$Q_{L=2500} \downarrow$} & \multicolumn{1}{l}{$\max{Q} \downarrow$} & \multicolumn{1}{l}{$L_{\max Q} \uparrow$} & \multicolumn{1}{l}{$\text{\#Inc}_{\max Q} \downarrow$} & \multicolumn{1}{l}{$\frac{Q_{L=2500}}{2500} \downarrow$} \\ \midrule
Random      & $H(s)$        & $2143\pm 14.3$                              & $2500$                                   & $2880\pm 13.8$                            & $1464 \pm 4.5$                                         & $85.72\%$                                                \\
Random      & $\log_2(|s|)$ & $1932\pm 3.7$                               & $2500$                                   & $\bm{3050\pm 1.3}$                             & $1360 \pm 2.6$                                         & $\bm{77.28\%}$                                                \\
Uncertainty & $H(s)$        & $2500\pm 0.0$                               & $2500$                                   & $2500\pm 0.0$                             & $\bm{0 \pm 0.0}$                                            & $100\%$                                                  \\
Uncertainty & $\log_2(|s|)$ & $\bm{1952\pm 1.0}$                               & $2500$                                   & $3032\pm 1.0$                             & $1372\pm 1.0$                                          & $78.08\%$                                                \\ \bottomrule
\end{tabular}
\end{table}

\begin{table}
\caption{Performance of methods over SVHN dataset using a pretrained predictor.}
\label{tab:svhn}
\centering
\begin{tabular}{@{}llrrrrr@{}}
\toprule
\cmidrule(l){3-7}
            &               & \multicolumn{5}{c}{SVHN}                                                                                                                                                                                                                               \\ \midrule
AL method   & Cost function & \multicolumn{1}{l}{$Q_{L=2500} \downarrow$} & \multicolumn{1}{l}{$\max{Q} \downarrow$} & \multicolumn{1}{l}{$L_{\max Q} \uparrow$} & \multicolumn{1}{l}{$\text{\#Inc}_{\max Q} \downarrow$} & \multicolumn{1}{l}{$\frac{Q_{L=2500}}{2500} \downarrow$} \\ \midrule
Random      & $H(s)$        & $438\pm 13.1$                               & $2208\pm 9.0$                            & $6000 \pm 0.4$                            & $895 \pm 11.4$                                         & $17.50\%$                                                \\
Random      & $\log_2(|s|)$ & $\bm{348 \pm 0.0}$                               & $2188 \pm 1.6$                           & $6000 \pm 0.0$                            & $1029 \pm 1.1$                                         & $\bm{13.92\%}$                                                \\
Uncertainty & $H(s)$        & $1262\pm 0.0$                               & $2199 \pm 0.0$                           & $6000 \pm 0.0$                            & $\bm{470 \pm 0.0}$                                          & $50.48\%$                                                \\
Uncertainty & $\log_2(|s|)$ & $\bm{348\pm 0.0}$                                & $\bm{2185 \pm 0.5}$                           & $6000 \pm 0.0$                            & $1029\pm 0.5$                                          & $\bm{13.92\%}$                                                \\ \bottomrule
\end{tabular}
\end{table}

\begin{table}
\caption{Performance of methods over MNIST dataset starting from scratch.}
\label{tab:mnist}
\centering
\begin{tabular}{@{}llrrrrr@{}}
\toprule
\cmidrule(l){3-7}
            &               & \multicolumn{5}{c}{MNIST}                                                                                                                                                                                                                              \\ \midrule
AL method   & Cost function & \multicolumn{1}{l}{$Q_{L=2500} \downarrow$} & \multicolumn{1}{l}{$\max{Q} \downarrow$} & \multicolumn{1}{l}{$L_{\max Q} \uparrow$} & \multicolumn{1}{l}{$\text{\#Inc}_{\max Q} \downarrow$} & \multicolumn{1}{l}{$\frac{Q_{L=2500}}{2500} \downarrow$} \\ \midrule
Random      & $H(s)$        & $740 \pm 46.7$                              & $2500$                                   & $5234\pm 52.6$                            & $1559\pm 21.8$            & $29.63\%$                                                \\
Random      & $\log_2(|s|)$ & $704 \pm 53.5$                              & $2500$                                   & $4971 \pm 15.9$                           & $1683\pm 2.8$                                          & $28.17\%$                                                \\
Uncertainty & $H(s)$        & $1198\pm 26.8$                              & $2500$                                   & $\bm{5716\pm 348}$                             & $\bm{611\pm 246}$                                           & $47.92\%$                                                \\
Uncertainty & $\log_2(|s|)$ & $\bm{699\pm 46.0}$                               & $2500$                                   & $4978 \pm 13.7$                           & $1683\pm 1.9$                                          & $\bm{27.94\%}$                                                \\ \bottomrule
\end{tabular}
\end{table}

\begin{table}
\caption{Performance of methods over FMNIST dataset starting from scratch.}
\label{tab:fmnist}
\centering
\begin{tabular}{@{}llrrrrr@{}}
\cmidrule(l){3-7}
            &               & \multicolumn{5}{c}{FMNIST}                                                                                                                                                                                                                             \\ \midrule
AL method   & Cost function & \multicolumn{1}{l}{$Q_{L=2500} \downarrow$} & \multicolumn{1}{l}{$\max{Q} \downarrow$} & \multicolumn{1}{l}{$L_{\max Q} \uparrow$} & \multicolumn{1}{l}{$\text{\#Inc}_{\max Q} \downarrow$} & \multicolumn{1}{l}{$\frac{Q_{L=2500}}{2500} \downarrow$} \\ \midrule
Random      & $H(s)$        & $818\pm 17.6$                               & $2500$                                   & $\bm{5039\pm 31.8}$                            & $1452\pm 19.8$                                         & $32.73\%$                                                \\
Random      & $\log_2(|s|)$ & $\bm{626\pm 66.6}$                               & $2500$                                   & $4733 \pm 161$                            & $1719\pm 17.2$                                         & $\bm{25.05\%}$                                                \\
Uncertainty & $H(s)$        & $1668 \pm 23.8$                             & $2500$                                   & $5006 \pm 107$                            & $\bm{248 \pm 84.7}$                                         & $66.7\%$                                                 \\
Uncertainty & $\log_2(|s|)$ & $630 \pm 71.1$                              & $2500$                                   & $4755 \pm 175$                            & $1718 \pm 19.4$\$                                      & $25.2\%$                                                 \\ \bottomrule
\end{tabular}
\end{table}

\section{Full results}
\subsection{Synthetic}
The numerical results for the synthetic experiments with the three synthetic datasets described in Section~\ref{sec:details} are presented in Table~\ref{tab:synthetic}.

\subsection{Real}
The numerical results for the experiments over real datasets as described in Section~\ref{sec:detailsreal} are presented in Tables~\ref{tab:cifar10},~\ref{tab:svhn},~\ref{tab:mnist} and~\ref{tab:fmnist}.

\end{document}